\RequirePackage{fix-cm}
\documentclass[twocolumn]{svjour3}          
\smartqed  
\usepackage{graphicx}
\usepackage[cmex10]{amsmath}
\usepackage{algorithmic}
\usepackage{algorithm2e}
\usepackage{multirow}

\usepackage{listings}

\begin{document}


\title{Optimizing L1 Cache for Embedded Systems through Grammatical Evolution}


\author{Josefa D\'{i}az \'{A}lvarez \and
        J. Manuel Colmenar \and 
        Jos\'{e} L. Risco-Mart\'{i}n \and
        Juan Lanchares \and
        Oscar Garnica
}


\institute{Josefa D\'{i}az \'{A}lvarez \at
              Centro Universitario de M\'{e}rida, Universidad de Extremadura, 06800 M\'{e}rida, Spain \\
              Tel.: +34-924-387268\\
              \email{mjdiaz@unex.es}           
           \and
	      J. Manuel Colmenar \at
	      Dept. of Computer Science and Statistics, Universidad Rey Juan Carlos, 28933 M\'{o}stoles, Spain \\
	      \email{josemanuel.colmenar@urjc.es}\\
           \and
	      Jos\'{e} L. Risco-Mart\'{i}n \and Juan Lanchares \and \'{O}scar Garnica \at
	      Dept. of Computer Architecture and Automation, Universidad Complutense de Madrid, 28040 Madrid, Spain\\
	      \email{\{jlrisco,julandan,ogarnica\}@dacya.ucm.es}\\
}

\date{Received: date / Accepted: date}

\maketitle
\begin{abstract}
Nowadays, embedded systems are provided with cache memories that are large enough to influence in both performance and energy consumption as never occurred before in this kind of systems. In addition, the cache memory system has been identified as a component that improves those metrics by adapting its configuration according to the memory access patterns of the applications being run. However, given that cache memories have many parameters which may be set to a high number of different values, designers face to a wide and time-consuming exploration space.
 
In this paper we propose an optimization framework based on Grammatical Evolution (GE) which is able to efficiently find the best cache configurations for a given set of benchmark applications. This metaheuristic allows an important reduction of the optimization runtime obtaining good results in a low number of generations. Besides, this reduction is also increased due to the efficient storage of evaluated caches. Moreover, we selected GE because the plasticity of the grammar eases the creation of phenotypes that form the call to the cache simulator required for the evaluation of the different configurations.

Experimental results for the Mediabench suite show that our proposal is able to find cache configurations that obtain an average improvement of $62\%$ versus a real world baseline configuration.

\end{abstract}

\section{Introduction}
Embedded computing systems have experienced a great development in last decades, becoming one of the crucial driving forces in technology. Portable devices such as smartphones, digital cameras, GPS, etc. are constantly incorporating new functionalities. Most of these new features are devoted to support new services and multimedia applications, despite that embedded systems have limited resources. Portable systems, for instance, run on batteries, which are limited in capacity and size, because of design constraints. Moreover, they have a limited cooling ability and, hence, a low energy consumption is a main requirement. Besides, portable systems usually execute multimedia applications, which require high performance and, therefore, are energy intensive tasks. As a consequence, one of the main workforces of embedded systems designers is the search for the right balance between increasing performance and reducing energy consumption at a low cost.

Previous studies have identified on-chip cache memory as one of the most energy consuming components, ranging between $20\%$ and $30\%$ of the total power of the chip in embedded processors~\cite{Varma05}. In addition, the design of the cache memory has a high influence on energy consumption and performance, because of the different hardware complexity that it implies. For instance, a direct mapped cache design presents the fastest hit times, but requires higher cache size to obtain comparable performance than an $N$-way set associative cache. On the other hand, set associative caches present higher energy consumption given that they access to a set of $N$ tags on each operation~\cite{Hennessy2011}.

The cache memory behavior is not only conditioned by structural parameters (capacity, block size, associativity, etc.), but also by parameters such as algorithms for search, prefetch and replacement, write policies, etc. All these features form the so-called cache configuration. Finding optimal values for these parameters will bring us the best performance and energy consumption and consequently, the best cache configuration.

Besides, applications have different behavior and hence have distinct cache requirements to meet the specific objectives of energy consumption and performance. Thus, we need to take into account the behavior of the target application in order to find the best cache configuration that improves performance or energy consumption.

In summary, an optimization scheme to determine the best cache configuration should: (1) travel along the design space of cache configurations, which is defined as the set of all combinations of values for all cache parameters; (2) take into account the energy and performance of target applications for each candidate cache configuration, because of different memory access patterns. 

The evaluation of a cache configuration requires a cache simulator to process a trace of the cache accesses from a previous execution of the target program. This process will allow to calculate the execution time and energy consumption for the application under study. This is a slow process because program traces usually record billions of memory operations, which spends tens of seconds of simulator runtime. Hence an exhaustive search on this design space will take an unaffordable amount of time given the high number of cache configurations that can be defined.

On the other hand, heuristic techniques reduce the number of evaluations while performing the search towards the optimal and, therefore, fit well in this kind of problems. Therefore, we present in this paper an optimization scheme based on Grammatical Evolution (GE)~\cite{Dempsey2009} that obtains, using program traces, the optimized cache configuration in terms of execution time and energy consumption for a given target application. Compared with the exhaustive approach, the execution time is highly reduced due to two reasons: (1) the metaheuristic algorithm converges even with a short number of generations and population size; (2) we have added a hashed map to store the objective values of each evaluated cache. In addition, the plasticity of the grammar in GE help us to generate the parameters that the cache simulator call requires for each evaluation, making more easy to communicate it with the optimization algorithm.

In order to test our proposal, we have run experiments that find the best cache configuration for a set of multimedia applications taken from the Mediabench benchmarks ~\cite{mediabench}, since they are representative for image, audio and video processing, which are typical applications of embedded systems. We have assumed in these experiments a hardware architecture based on the ARM9 processors family~\cite{ARM9}, broadly used in multimedia embedded devices. As will be shown later, the average reduction of execution time and energy consumption of the optimized cache configurations is $75\%$ and $96\%$ respectively in relation to a baseline cache. In addition, the optimization runtime is affordable and we have estimated an average reduction of $94\%$ in relation to a GE implementation with no storage of evaluations.

The rest of the paper is organized as follows. Section \ref{sec:Cache} describes the main issues about cache design and defines the search space of the problem. Section \ref{sec:related} reviews the related work on the cache optimization techniques. Section~\ref{sec:model} describe both performance and energy models adopted in this work to compute the goodness of the candidate solutions. Section ~\ref{sec:Methodology} describes the whole optimization framework, detailing the off-line processes as well as the Grammatical Evolution we manage. Then, Section \ref{sec:Experiments} analyzes our experimental results. Finally, Section ~\ref{sec:Conclusions} draws conclusions and future work.


\section{Cache design and search space}
\label{sec:Cache}
In order to clarify the scope of the problem under study, we devote this section to introduce some cache design concepts and related issues to be addressed. These concepts will also help us determine the size of the search space.

Firstly, it must be noticed that there exist diferent design features, also named design parameters, that need to be tuned in order to decide a cache configuration. The values given to those parameters influence on the cache behavior in both execution time and energy consumption. However, those parameters may be different depending on the processor where the cache will be included.

In this regard, we first have to decide which type of cache to select for this work. Taking into account that we focus our approach on embedded systems, we have selected the cache of the ARM9 processors family~\cite{ARM9}.

ARM processors, because of their low power consumption and cost, have a prevalent position in the market of portable devices and they are widespread on multimedia embedded devices. In fact, the Apple A7, A8 and A8X processors included in the popular iPhone 5S, iPhone 6 and iPad Air 2 do implement the ARMv8-A instruction set.

In addition, the ARM9 processors family is present in devices such as game consoles (GP32, Nintendo DS), calculators (HP 49/50), mobile phones (HTC TyTN, K and W series of Sony Ericsson) and car GPS (Samsung S3C2410), for example. 

Processors belonging to this family allow the installation of a cache memory with sizes between 4 KB and 128 KB. For instance, the ARM920T, ARM922T and ARM940T processors (ARM9TDMI family) implement a separated 16KB, 8KB and 4KB instruction and data cache, respectively. Although revisions on ARM9E family allow sizes up to 1MB, such as the ARM946E-S processor. 

Given that the separation into data and instruction caches is common in the more recent ARM9 processors, we do consider this two-caches approach in our work. 

Hence, the cache parameters that can be tuned in the instruction and data caches of this processor are the following: cache size, line size, replacement algorithm, associativity and prefetch algorithm for both the instruction and data caches, and also write policy for data cache. Here, cache size is the memory cache capacity in bytes, block size represents the amount of data read or written in each cache access, replacement algorithm is in charge of taking the decision to evict a block from cache memory, and prefetch algorithm decides how blocks are carried to the cache memory. Finally, the write policy decides when data stored in the cache shall be written to main memory, and it only affects the data cache.

\begin{figure}[ht]
\centering
 \includegraphics[width=0.96\columnwidth]{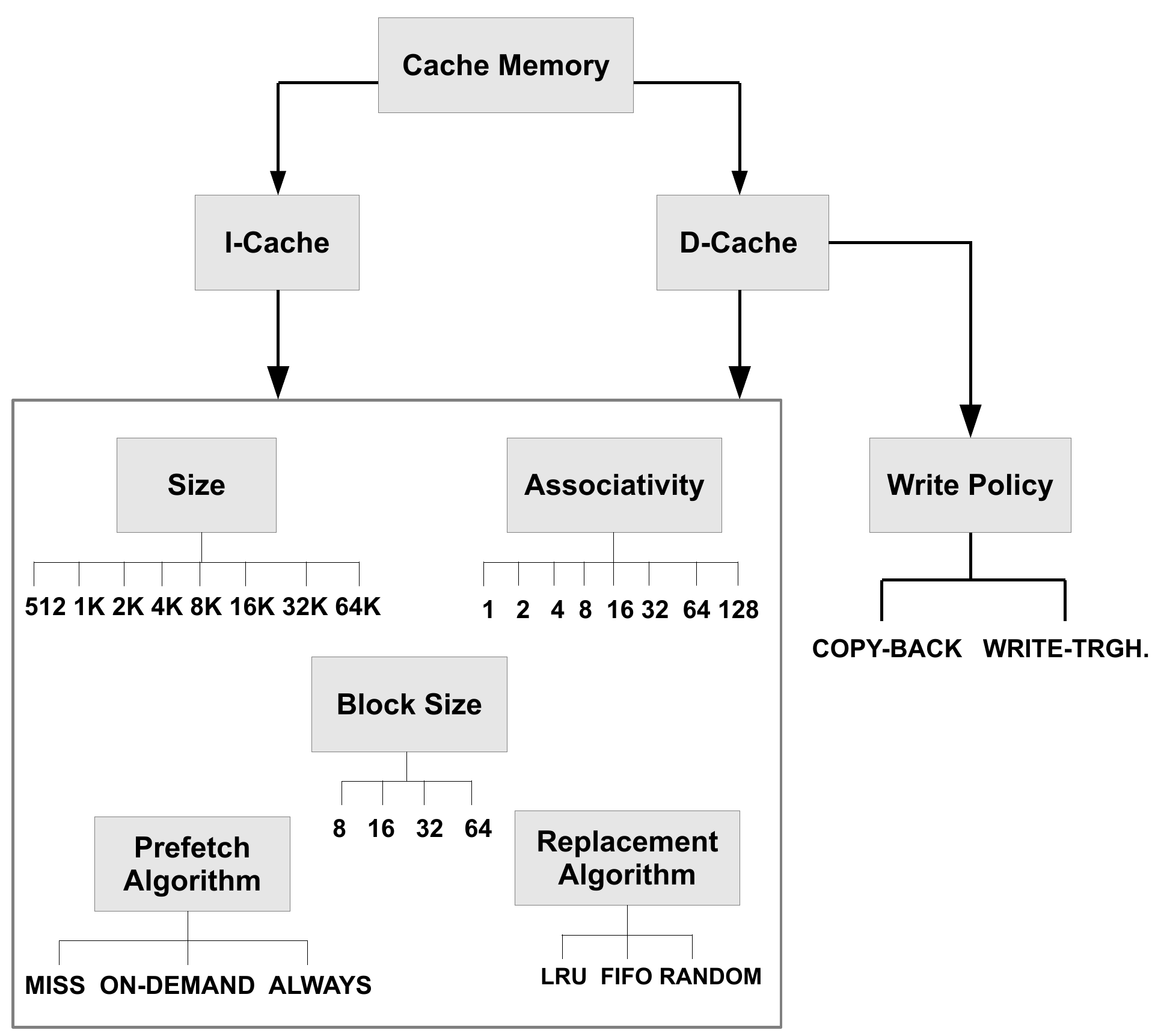}
 \caption{Cache configuration parameters. Both instruction and data caches must be customized with available values.}
  \label{figure:arboles}
\end{figure}

In brief, the problem we address consists on selecting the values for the cache parameters that optimize the execution time and energy consumption of a given application. Therefore, the search space is formed by all the combinations of different values for each one of the cache parameters. Thus, in our target system, a cache configuration is formed by 11 parameters: 5 for the instructions cache and 6 for the data cache. According to the typical cache configurations of embedded systems, we selected a range of permitted values for each parameter, which are shown in \figurename~\ref{figure:arboles}.

For instructions cache (I-Cache), memory size may take 8 different values ranging between 512 bytes and 64 KB, block size can take 4 different values, 3 replacement algorithms and 3 prefetch algorithms may be selected, and 8 associativities are available. This results in $2304$ possible configurations for instructions cache.

Data cache (D-Cache) have the same parameters as instructions cache, but adding the 2 possible write policies. Hence, $4608$ possible data cache configurations can be selected.

Thus, considering all the possible combinations of both caches, the size of the search space is $2304 \times 4608 = 10616832$ configurations. However, some of the constraints are related to infeasible parameters combinations. For example, a cache configuration of 512B, 64-ways set associative with 32 bytes block size is not feasible. Then, if this combination is selected for instruction or data caches, it shall be invalidated by any optimization algorithm.


\section{Related work}
\label{sec:related}
It is well proven that the cache memory is a component that plays a crucial role to improve performance and energy consumption. In recent decades, many researchers have focused their work on improving different characteristics of the cache memory. Several optimization approaches have been addressed to increase performance and/or reduce energy consumption, develop simulation and evaluation tools, etc. Next, we analyze some of the works related to our proposal that we have separted into two different cathegories: (1) works that deal with cache reconfiguration, and (2) works based on evolutionary techniques.

\subsection{Cache reconfiguration}

One of the main goals for years has been to configure the cache memory according to the running workload, in order to mainly improve performance and/or energy consumption. This requires both hardware support for cache reconfiguration, and algorithms to decide the proper values of the parameters that obtain the best performance taking into account the running application. In this way, many research works have analyzed cache memory parameters and proposed new reconfiguration techniques from different points of view.

Naz et al.~\cite{Naz-2005} proposed a design with split data caches for embedded systems. The data cache was split in scalar and array caches aiming to take advantage of temporal and spatial localities respectively. Their optimization was based on dealing with predefined cache configurations, where the values of the parameters were fixed before the optimization. Hence, the search space in this case is small compared with a space where all the parameter values are taken into account.

Chen and Zou~\cite{Chen-2007} presented an efficient reconfiguration management algorithm for embedded systems. The cache configuration automatically changed after identifying a phase change. The search space was composed by these parameters: cache size, block size and associativity. From our point of view, this work do not consider enough number of cache parameters.

Gordon-Ross et al.~\cite{Gordon-2008} applied a method with an off-line phase classification and an on-line predictor phase. Phase classification breaks applications execution into fixed size intervals grouped according to its similar behavior pattern. The phase predictor decides the cache configuration for the next interval. Exploration space was defined by cache size, block size and associativity, which means that only three parameters were considered to define the search space. In this case, given the on-line prediction phase, a low number of parameters was required.

Wang et al.~\cite{Wang-2012} proposed dynamic cache reconfiguration for real time embedded systems. They minimize energy consumption performing a static analysis at runtime. This proposal is based in previous works~\cite{Wang-2009,Wang-2011} where they addressed an unified two-level cache hierarchy and multi-level cache hierarchy respectively. A combination of static analysis and dynamic tuning of cache parameters, without time overhead were performed in both cases. However, a few number of parameters were optimized: cache size (1, 2 or 4KB), line-size (16, 32 or 64 bytes) and associativity (1-way, 2-way or 4-way). 

New hardware technologies and core-based processor technologies, such as those proposed, for example, by ARM~\cite{ARMreconf2014}, allow changing the cache configuration for each application. Changes affect the main parameters: capacity, block size and associativity. However, an efficient algorithm is needed to determine optimal values for each application.

\subsection{Evolutionary Techniques}

Evolutionary techniques have been widely applied in design optimization, both for hardware and software optimization with different objectives.

The creators of the Grammatical Evolution technique published a work where a caching algorithm was automatically generated ~\cite{ONeill99automaticgeneration}. However, they only considered the generation of the caching algorithm and kept fixed some other parameters like cache size, for instance. Again, many parameters were missed.

Palesi and Givargis~\cite{Palesi-2002} presented a  multi-objective optimization approach to explore the design space of a system-on-a-chip (SoC) architecture to find the pareto optimal configurations of the system. They performed the optimization through a genetic algorithm where the evaluation is done directly in the SoC. However, the number of parameters they study is lower than ours, and their approach is not escalable to current multimedia applications given the low performance of the SoC.

Filho et al. ~\cite{Filho-2008} presented an approach based on the NSGA-II algorithm to evaluate cache configurations on a second level cache in order to improve energy consumption and performance, optimizing cache size, line size and associativity. However, again, the number of parameters is lower than in our proposal and, therefore, the search space is smaller.

Dani et al.~\cite{Dani2012} applied a genetic algorithm to find an optimal cache configuration for a chip multiprocessor (CMP) system. They deal with a complex system like a CMP. However, they do not manage a higher number of cache parameters and, given that they encode the configuration of both the chip and the cache into a chromosome, the decodification process is complex. This issue makes difficult the application of the same method to a different microarchitecture.

Risco et al.~\cite{RiscoMartin2014109} presented a framework based on Grammatical Evolution  to design custom dynamic memory managers for multimedia applications in embedded systems. They optimized memory accesses, memory usage and energy consumption. However, they do not deal with cache parameters but optimizing the operating system memory manager instead of dealing with cache design parameters.

Finally, performing a deep search in the literature, and specially in one of the main hubs of Grammatical Evolution publications \cite{GEpubs2015}, we have not found any recent work which applied this technique to optimize the cache design of a microarchitecture.

\subsection{Summary} 

As seen, the approaches that make use of reconfiguration require extra hardware complexity in the design of the memory subsystem. Also, in the majority of the cases, this complexity adds an overhead in execution time. In our work we propose a different approach focused on finding the best cache configuration after exploring the search space of cache designs. This strategy avoids the cache reconfiguration which does not add hardware complexity to the cache memory design.

In addition, previous works define their design space by cache size, block size and associativity of either instruction cache or data cache, but we have not found any proposal that considers both caches at the same time on embedded systems. Our motivation is different since we consider the optimization of instruction and data caches together, and we selected all the parameters that can be tunned: cache size, block size, associativity, replacement algorithm and prefetch algorithm for both caches; and write policy for data cache. Therefore, we explore a wider search space.
 
Moreover, to the best of our knowledge, none of the previous works considers the optimization of as many parameters as we tackle in our proposal. Regarding the works using evolutionary techniques, most of the cited papers have focused their space exploration on typical structural cache parameters such as cache size, line size and associativity and, in addition, they usually deal with short ranges for parameter values. One of the most used technique in this kind of works, the genetic algorithm, presents the drawback of the codification process which has to be customized for every different cache paradigm and simulator software.

On the other hand, the methodology we propose can be applied to any kind of cache and processor architectures. To this aim, the main and unique change in our framework will be the cache simulator and the grammar, as it will be shown later.


\section{Performance and energy models}
\label{sec:model}

In this research work we consider the execution time and energy consumption of a given cache configuration when running a benchmark application. As explained before, we deal with program traces that are processed by a cache simulator that accounts for the number of hits and misses of each execution, which represent the cache behavior.

As stated before, we selected the ARM9 processors family~\cite{ARM9} as our target embedded system where cache configuration will be optimized. Here, the L1 cache is divided into instructions cache and data cache, and an embedded DRAM acts as main memory.

In addition to the target architecture, we consider that instructions cache is read-only, and a main memory is available. The main memory features are displayed in Table \ref{Table:dram} and will be taken into account in the evluation of a cache configuration.

\begin{table}
\renewcommand{\arraystretch}{1.3}
\centering
\caption{Main memory (DRAM) features.}
\label{Table:dram}
\begin{tabular}{|c|c|}
\hline
Size & $64$ Mb \\
\hline
Access time & $3.9889 \times 10^{-9}$ sec. \\
\hline
Bandwidth & $6.7108864 \times 10^{9}$ bytes/sec. \\ 
\hline
Access power & $1.051$ W \\ 
\hline
\end{tabular}
\end{table}

Once the hardware features are set, we need a model able to translate the number of hits and misses of the benchmarks runs into execution times and energy consumption. To this aim, we chose the performance and energy models described in~\cite{Janapsatya-2006} for our optimization framework.

The authors of the paper proved, with a 100\% accuracy, that the models were equivalent to the results obtained from a cache simulator. Therefore, given this performance, and taking into account that the expressions of the models could be evaluated very fast, they fit very well into an optimization algorithm like the one we present.

Hence, considering these models, we do account the performance and energy model of the cache memory subsystem. The model of the rest of the components of the chip (CPU, main memory, etc.) is not taken into account given that our optimization scheme tackles only the cache subsystem. Next we briefly describe these cache models.

\subsection{Performance model}\label{sec:pe_model}

The cache performance model allows to obtain the execution time. This model is based on the number of hits and misses on the cache memory system and the time needed to solve them. We show in (\ref{eq:extime}) the equation applied to compute execution time. Each one of its components is described bellow, but a wider explanation can be found in~\cite{Janapsatya-2006}.

\begin{scriptsize}
\setlength{\arraycolsep}{0.0em}
\begin{eqnarray}
T&\,=\,& Icache_{access} \times Icache_{access\_time} + \nonumber \\ 
  & & Icache_{miss} \times DRAM_{access\_time} + \nonumber \\
  & & Icache_{miss} \times Icache_{line\_size} \times \frac{1}{DRAM\_bwidth} + \nonumber\\
  & & Dcache_{access} \times Dcache_{access\_time} + \nonumber \\
  & & Dcache_{miss} \times DRAM_{access\_time} + \nonumber \\
  & & Dcache_{miss} \times Dcache_{line\_size} \times \frac{1}{DRAM\_bwidth} \label{eq:extime}
\end{eqnarray}
\end{scriptsize}

Each term in the equation represents:
\begin{itemize}
 \item $Icache_{access}$ and $Dcache_{access}$ are the number of cache memory accesses to the instruction and data caches, respectively.
 \item $Icache_{miss}$ and $Dcache_{miss}$ are the number of cache misses. For each one, data must be copied from the main memory.
 \item $Icache_{access\_time}$ and $Dcache_{access\_time}$ represent the time needed for each access to the instruction and data caches, respectively.
 \item $DRAM_{access\_time}$ is the main memory latency time.
 \item $Icache_{line\_size}$ and $Dcache_{line\_size}$ correspond to the line size (block size) for instruction and data caches, respectively. 
 \item $DRAM\_{bwidth}$ is the transfer capacity of the DRAM.
\end{itemize}

Therefore, the first and fourth parts in this equation represented by $Icache_{access} \times Icache_{access\_ti\-me}$ and $Dcache\-_{access} \times Dca\-che_{access\_ti\-me}$ compute the total time needed to solve all instruction and data cache accesses, respectively. $Ica\-che_{miss} \times DRAM_{access\_ti\-me}$ and $Dca\-che_{miss} \times DRAM_{access\_ti\-me}$ are the total time spent by main memory accesses in response to instruction and data cache misses. The third and sixth parts, specified by $Ica\-che_{miss} \times Ica\-che_{li\-ne\_si\-ze} \times \frac{1}{DRAM\_bwidth}$ and $Dca\-che_{miss} \times Dca\-che_{li\-ne\_si\-ze} \times \frac{1}{DRAM\_bwidth}$  calculate the total time needed to fill in an instruction and data caches line when a miss happens, respectively.

\subsection{Energy model}

The energy model is defined in (\ref{eq:energy1}), and also detailed in ~\cite{Janapsatya-2006}. This equation is used to determine the dynamic energy consumption for a cache configuration.

\begin{scriptsize}
\setlength{\arraycolsep}{0.0em}
\begin{eqnarray}
E &\,=\,& Total_{time} \times CPU_{power} + \nonumber \\ 
& & Icache_{access} \times Icache_{access\_energy} +  \nonumber \\ 
& & Dcache_{access} \times Dcache_{access\_energy} + \nonumber \\
& & Icache_{miss} \times Icache_{access\_energy} \times Icache_{line\_size} + \nonumber \\
& & Dcache_{miss} \times Dcache_{access\_energy} \times Dcache_{line\_size} + \nonumber \\
& & Icache_{miss} \times DRAM_{access\_power} + \nonumber \\
& & (\small{DRAM_{access\_time}} + \small{Icache_{line\_size}} \times \frac{1}{\small{DRAM_{bwidth}}}) + \nonumber \\
& & Dcache_{miss} \times DRAM_{access\_power} + \nonumber \\
& & (\scriptsize{DRAM_{access\_time}} + \scriptsize{Dcache_{line\_size}} \times \frac{1}{\tiny{DRAM_{bwidth}}}) 
\label{eq:energy1}
\end{eqnarray}
\end{scriptsize}

Most of the components of the equation have been briefly described above. The rest of the components are the following:

\begin{itemize}
 \item $DRAM_{access\_power}$ represents the power consumption for each DRAM access.
 \item $Icache_{access\_energy}$  and $Dcache_{access\_energy}$ correspond to energy consumption for each instruction and data caches access, respectively.
\end{itemize}

The $Icache_{access} \times Icache_{access\_energy}$ and $Dcache\-_{access} \times Dcache_{access\_energy}$ terms calculate the energy consumption due to instruction and data caches, respectively. $Icache_{miss} \times Icache_{access\_energy} \times Icache\-_{line\-\_size}$ and $Dcache_{miss} \times Dcache_{access\_energy} \times Dcache\-_{line\-\_size}$ is the energy cost of writing data to the instruction and data caches when misses occur. The two last terms calculate the energy cost of the DRAM when responding to cache misses.

In our approach, we have removed the first term of the energy equation $Total_{time} \times CPU_{power}$ because of three reasons: (1) the term $CPU_{power}$ is constant and the term $Total_{time} $ corresponds to the total time of the system, not only the cache subsystem, (2) it represents the amount of energy consumed by the CPU and we are optimizing just the performance and energy consumed by the memory subsystem, and (3) this term could be large enough to hide the differences because of memory cache operations. Thus, the energy equation is reduced to:

\begin{footnotesize}
\setlength{\arraycolsep}{0.0em}
\begin{eqnarray}
E &\,=\,& Icache_{access} \times Icache_{access\_energy} +  \nonumber \\ 
& & Dcache_{access} \times Dcache_{access\_energy} + \nonumber \\
& & Icache_{miss} \times Icache_{access\_energy} \times Icache_{line\_size} + \nonumber \\
& & Dcache_{miss} \times Dcache_{access\_energy} \times Dcache_{line\_size} + \nonumber \\
& & Icache_{miss} \times DRAM_{access\_power} + \nonumber \\
& & (\small{DRAM_{access\_time}} + \small{Icache_{line\_size}} \times \frac{1}{\small{DRAM_{bwidth}}}) + \nonumber \\
& & Dcache_{miss} \times DRAM_{access\_power} + \nonumber \\
& & (\small{DRAM_{access\_time}} + \small{Dcache_{line\_size}} \times \frac{1}{\small{DRAM_{bwidth}}}) 
\label{eq:energy}
\end{eqnarray}
\end{footnotesize}

All the equations use seconds, watts, joules, bytes and bytes/sec as units to measure time, power, energy, cache line size and bandwidth, respectively.

Next, we describe the whole process of optimization showing where these models are applied.


\section{Optimization methodology}
\label{sec:Methodology}

As explained before, in this work we propose an approach to identify the best cache configuration for a given set of applications. We define the best cache configuration as the one which minimizes the execution time and energy consumption for a given benchmark.

\begin{figure}[ht]
\centering
 \includegraphics[width=0.95\columnwidth]{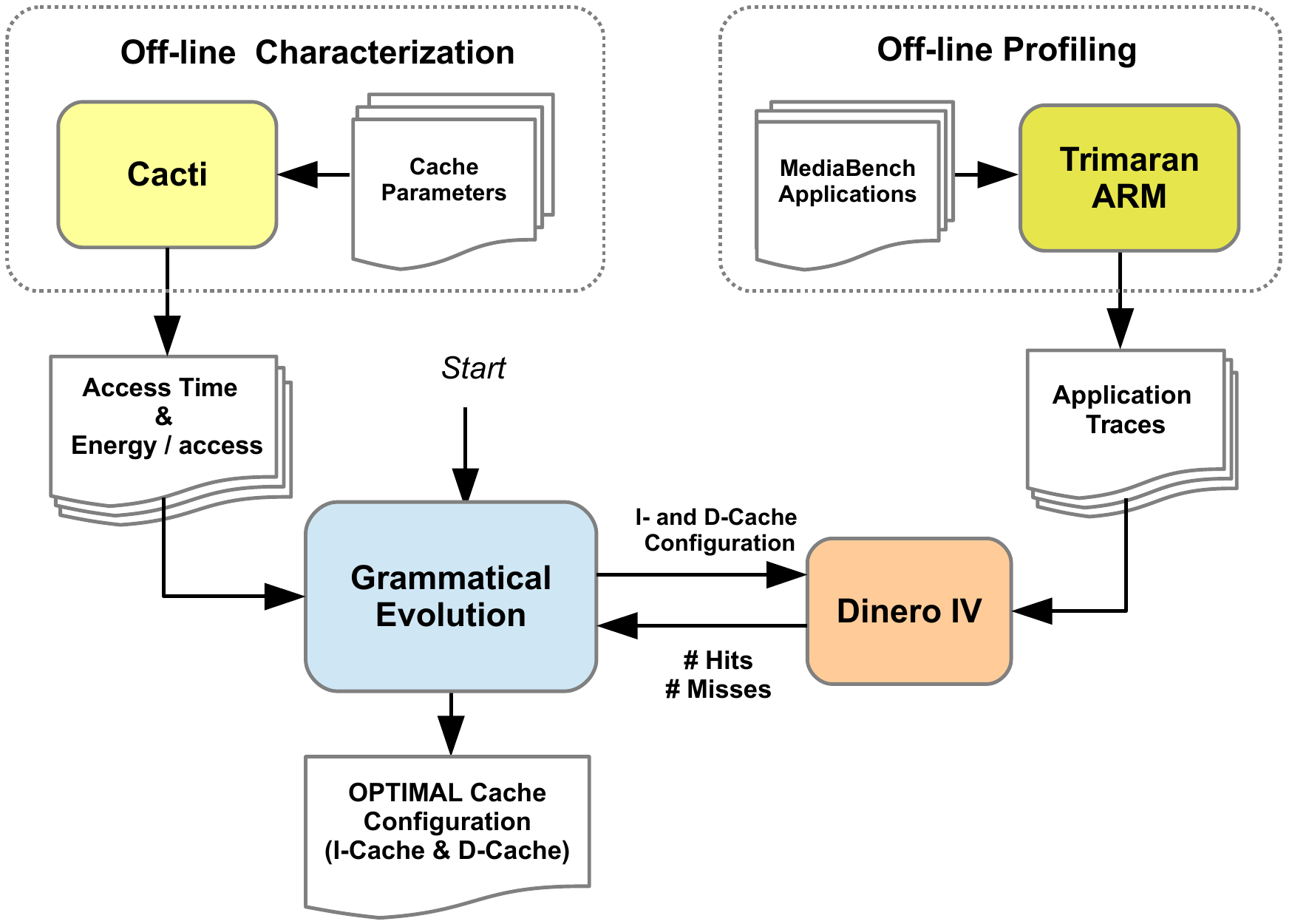}
 \caption{Processes involved in the cache configuration optimization.} 
\label{figure:method}
\end{figure}

With this aim in mind, in this section we describe the optimization framework that we propose to optimize the cache memory, in this case, of multimedia embedded systems. \figurename~\ref{figure:method} presents an overview of the whole process.

We divided our optimization process into three different stages. Firstly, two processes are run just once before the optimization. These are called off-line processes and perform: (1) cache characterization and (2) application profiling. The third stage is the optimization, which is formed by the evolutionary process running GE and calling the cache simulator. 

The optimization is performed by the Grammatical Evolution (GE) module, as seen in \figurename~\ref{figure:method}. Each time a configuration has to be evaluated, the GE calls Dinero IV, which is a trace-driven cache simulator~\cite{DineroIV}. It receives a cache configuration from GE and returns the number of cache hits and misses for the target application traces. These date are included in the models to evaluate the configuration. Next, we give more detail of this optimization flow.

\subsection{Off-line process}

Two tasks are run off-line: the characterization of the possible cache configurations, and the profiling of the target benchmarks. Both tasks are slow, but can be run just once, because their results are stored in data files that are read by the cache simulator and the GE module.

The characterization process, on the upper left part of \figurename~\ref{figure:method}, has the specific task of combining all the parameter values for cache configurations. That is, generates all the possible cache configurations and obtains their access time and energy consumption per access. Hence, this process provides the time and energy values used for the terms in equations (\ref{eq:extime}) and (\ref{eq:energy}). For this purpose we used Cacti~\cite{Mamidipaka2004} considering 32nm technology to compute the DRAM and cache access times and dynamic energy consumed for each cache access. Cacti is an analytical model widely applied to estimate energy and power consumption, performance and area of cache memories. This characterization was run only once, and we selected for this task the web tool provided by the Cacti organization. The most important parameters required by Cacti are those mainly related to hardware: cache size, line size and associativity, and there is no distinction between instructions and data caches. Hence, this is a slow but affordable task.

The profiling process, on the upper right part of \figurename~\ref{figure:method}, is in charge of the generation of the traces from the target applications. Again, this is a process that has to be run just once. The profiling takes each benchmark application and obtains the sequence of cache hits and misses from a run, generating a memory trace file. This process is carried out by the open-source compiler and performance monitoring infrastructure called Trimaran~\cite{trimaran}, which provides the resources required to obtain accurate application traces. Given that the ARM processors are not directly addressed by Trimaran, we have modified it by incorporating code from the architectural simulator SimpleScalar~\cite{simplescalar}, which is able to model a wide set of different architectures, being the ARM among them.

Besides, it has to be mentioned that input and output data of all the benchmarks under study are represented in the trace file. Therefore, in order to capture the mean behavior of each application, we will select the recommended inputs indicated in the documentation of each benchmark.

\subsection{Grammatical Evolution}
\label{GE-phase}

The characterization of the cache ends up with two kind of results. On the one hand, the time and energy required per access is obtained for each cache configuration and, on the other hand, the traces with all the memory operations of each benchmark application have been obtained. Therefore, our framework is ready to evaluate the performance of any candidate cache configuration, which will be generated in the optimization module, based on Grammatical Evolution.



Grammatical Evolution~\cite{ONeill2001,ONeill2003,ONeill2008,Dempsey2009} is an evolutionary technique that is able to represent individuals by means of grammars. GE is a form of Genetic Programming (GP)~\cite{Koza2003} which evolves programs that are evaluated according their fitness values. In this regard, GE represents each program, that is, each individual, through an expression generated by a grammar defined by the researcher.

GE performs evolutionary process on variable-length binary strings, which are mapped to generate a program by selecting production rules in a Backus-Naur form (BNF) grammar definition. Given that strings are similar to chromosomes, GE allows using genetic operators like selection, crossover and mutation.
Problems such as financial modeling~\cite{Brabazon2006}, game strategies~\cite{Galvan-2010}, dynamic game environment~\cite{Perez-2011}, 2D platforms games design~\cite{Shaker-2012}, 3D design~\cite{Byrne-2010} among others, have been successfully addressed by using GE.

So, the GE traverses the search space of cache configurations according to the grammar definition. For each individual, it calls the cache simulator to obtain the number of hits and misses for the target benchmarks. Then, using the access time and energy per access previously computed by Cacti, and the performance and energy models described in Section \ref{sec:model}, the fitness is computed. This process is repeated till the selected number of generations for GE is reached. As a result, an optimized cache configuration is obtained for each one of the benchmark programs.

As seen in Section \ref{sec:Cache}, the search space of this problem is defined by the set of cache configurations. However, GE works with individuals that represent those configurations under a proper codification. In fact, GE uses a linear genome rather than a tree-based structure, as in GP. Linear genome facilitates the application of genetic operators. Here, the chromosome of each individual consists of a string of integer values corresponding to codons. Each codon stores a value of 8 bits, hence forming the genotype. The genotype is decoded in a process called mapping, where a grammar is used to translate the codons into the phenotype of the individual.


One individual represents a particular cache configuration that must be evaluated in order to measure its execution time and energy consumption. In this regard, we have decided to consider the same weight for both features of the cache. Therefore, for a given cache configuration $c_i$, the evaluation process is conducted by the fitness function $f$ defined in (\ref{eq:fitness}).

\begin{footnotesize}
\begin{equation}
\centering
f(c_i)=0.5\times \frac{T(c_i)}{T(Baseline)}+0.5\times \frac{E(c_i)}{E(Baseline)}
 \label{eq:fitness}
\end{equation}
\end{footnotesize}

Each component of the fitness, that is, execution time and energy consumption, are normalized with respect to a baseline configuration. By normalizing to a baseline we are able to correctly scale and operate the units of different measures like time and energy. As seen in (\ref{eq:fitness}), both terms are weighted to $50\%$ seeking the best balance between the improvement of the performance and energy consumption. The best cache configuration in GE will be the one that minimizes the fitness value.


Given that the evaluation of a cache configuration involves a call to the cache simulator, we have designed a grammar that produces phenotypes which form the parameters of the call to the simulator. In fact, one of the benefits of GE is the ability to create complex phenotypes from a genotype formed by integer values.

\figurename~\ref{figure:grammar} shows, in BNF format, the grammar we have designed for the cache optimization under the previously defined search space. As stated before, the grammar produces the string of parameters that will be sent to the cache simulator. In fact, we have adapted the parameters to the requirements of the simulator. As seen in the figure, the replacement algorithm will take values like \texttt{l}, \texttt{f} or \texttt{r}, which correspond in the simulator, respectively, to \texttt{LRU}, \texttt{FIFO} and \texttt{RANDOM}. The same applies for prefetching and write policy.

According to the terminology, a grammar is described as a tuple \{\textit{N, T, S, P}\}, where \textit{N} represents the non-terminals symbols, \textit{T} is the set of terminals, \textit{S} represents the start symbol belonging to \textit{N}, and \textit{P} defines the set of production rules that map the values of \textit{N}  to \textit{T}. The ``$|$'' symbol separates the different choices within a single production rule. In our grammar, \textit{P} is formed by production rules from \texttt{I} to \texttt{VII}.

\begin{figure}[!t]
\begin{lstlisting}[basicstyle=\scriptsize\ttfamily,breaklines=true,frame=tb]
N = { <DineroParams>, <CacheSizeB>, <LineSizeB>, 
      <ReplAlg>, <Assoc>, <PrefAlg>, <WritePol> }
T = { 1, 2, 4, 8, 16, 32, 64, 128 512, 1024, 2048,
      4096, 8192, 16384, 32768, 65536, l, f, r, m,
      d, a, n }
S = <DineroParams>

P = { I    <DineroParams> ::= -l1-isize <CacheSizeB>
                              -l1-ibsize <LineSizeB>
                              -l1-irepl <ReplAlg>
                              -l1-iassoc <Assoc> 
                              -l1-ifetch <PrefAlg>
                              -l1-dsize <CacheSizeB>
                              -l1-dbsize <LineSizeB>
                              -l1-drepl <ReplAlg> 
                              -l1-dassoc <Assoc>
                              -l1-dfetch <PrefAlg>
                              -l1-dwback <WritePol>
                              
      II    <CacheSizeB> ::= 512 | 1024 | 2048 | 4096 
                           | 8192 | 16384 | 32768 
                           | 65536
      III   <LineSizeB> ::= 8 | 16 | 32 | 64
      IV    <ReplAlg> ::= l | f | r
      V     <Assoc> ::= 1 | 2 | 4 | 8 | 16 | 32 | 64 
                      | 128
      VI    <PrefAlg> ::= m | d | a
      VII    <WritePol> ::= a | n  }
\end{lstlisting}

 \caption{Grammar for cache configuration description.}
  \label{figure:grammar}
\end{figure}

\begin{figure}[ht]
\centering
 \includegraphics[width=0.99\columnwidth]{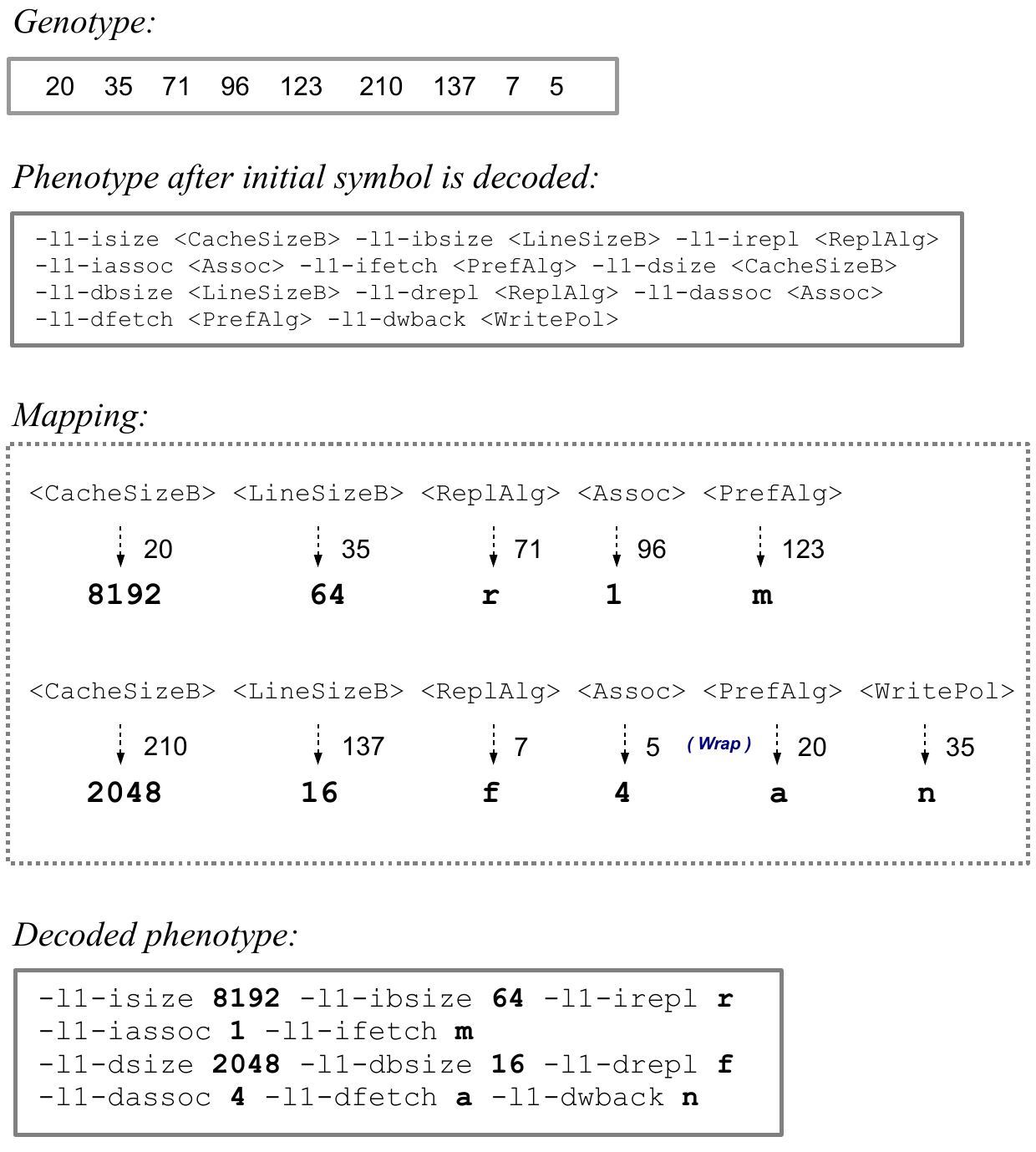}
	\caption{Decoding process in GE that, starting from the genotype above, obtains the phenotype by mapping through the proposed grammar.}
\label{fig:decoding}
\end{figure}

Next we show an example of the mapping process for the problem at hand using the grammar in \figurename~\ref{figure:grammar}. Let's consider the genotype shown in the upper part of \figurename~\ref{fig:decoding} as a candidate solution. This individual is a genotype that represents a cache configuration. However, it has to be decoded in order to obtain the values of each one of the 11 parameters of the cache configuration previously described.

The mapping or decodification process is performed by applying the modulus operation between the codon value and the number of options of each production rule corresponding to the non-terminal symbol being processed.

Then, considering the example individual, the decoding process is performed as shown in \figurename~\ref{fig:decoding}. It begins decoding the starting symbol, \texttt{<DineroParams>}, which is substituted by its production, that corresponds to the parameterized call to the cache simulator. Here, several non-terminal symbols are present, and the first one is \texttt{<CacheSizeB>}. Hence, the first gene of the genotype is selected, $20$. Given that the production for \texttt{<CacheSizeB>} (rule \texttt{II}) has $8$ different options, the selected value is option $4$ ($20\ MOD\ 8 = 4$), which corresponds to \texttt{8192}. Notice that the first option of the rule is linked to $0$, the second to $1$ and so on. Next, the second gene is read and its value, $35$, is used to decode the following non-terminal symbol, \texttt{<LineSizeB>}. The production rule for this symbol is \texttt{III}, and has $4$ options. Then, after the modulus operation ($35\ MOD\ 4 = 3$) \texttt{<LineSizeB>} is replaced by \texttt{64}. The following non-terminal symbol to be decoded is \texttt{<ReplAlg>}, whose production rule, \texttt{IV} has $3$ different options. Then, the codon value, $71$, is processed. Given that $71\ MOD\ 3 = 2$, the selected value is \texttt{r}. In an analogous way, the next codon, $96$ for \texttt{<Assoc>} is turned into $1$ given that $96\ MOD\ 8 = 0$. The last non-terminal corresponding to the instructions cache parameters, \texttt{<PrefAlg>}, is determined by the codon value $123$ which is replaced by \texttt{m}, once that $123\ MOD\ 3 = 0$. Next, the mapping process continues with the codon $210$ and the second \texttt{<CacheSizeB>} symbol, which maps $2048$ and corresponds to the data cache size. Then, $137$ maps \texttt{<LineSizeB>} to $16$, gene $7$ maps \texttt{<ReplAlg>} to \texttt{f} and $5$ maps \texttt{<Assoc>} to $4$.

At this point, all the genes have been used but the expression is still partially decoded. However, GE is able to work with variable length chromosomes and, if the decodification process reaches the end of the chromosome and there are still non-terminal symbols, GE goes to the first codon and allows the decodification to finish. This process is called \textit{wrapping}. Hence, in our example a \textit{wrap} takes place at this point. Therefore, \texttt{<PrefAlg>} is decoded with the first gene, $20$, obtaining \texttt{a}. Finally, decodification ends mapping \texttt{<WritePol>} to \texttt{n} with gene value $35$.

As a result, our genotype example returns the phenotype shown in the lower part of \figurename \ref{fig:decoding}, which represents a cache configuration where instructions cache size is $8192$, instructions cache block size is $64$, and so on.

Once decoded, the phenotype is used to call the cache simulator in order to evaluate the corresponding configuration on the target application, obtaining both the execution time and energy consumption according to the fitness defined in (\ref{eq:fitness}). At the end of the selected number of generations, the GE returns the cache configuration that obtains the best fitness value. 

As seen, the grammar determines both the concrete parameters to be optimized, as well as the communication with the cache simulator, which provides the fitness value. Therefore, changing the grammar will allow different optimization approaches: fix the value of some parameters, add new parameters, change the cache simulator, use variable length arguments in the cache simulator (for custom algorithms, for example), an so on. 

In this way, given both an invariable number (and type) of parameters and a cache simulator, the grammar forces the evolution of a fixed structure, and GE acts optimising the parameters in this structure. Hence, this problem could be tackled with a fixed-length algorithm like a GA, PSO and so on. However, in this kind of approaches, any change in the number or type of parameters, or even in the simulator, will require the modification of an important part of the code, mainly in the decodification and fitness evaluation modules. Hence, our approach based on GE provides a great flexibility as well as extensibility features because those changes could be incorporated with a simple change of the grammar.

The GE algorithm was developed in Java by our research team, and is publicly available as the JECO library~\cite{JECO}.


\section{Experimental results}
\label{sec:Experiments}

In order to test our experimental framework, we selected twelve benchmarks from the Mediabench suite~\cite{mediabench}, which are representative of multimedia applications. Next, we describe the features of these benchmarks. The full description can be found on the Mediabench website.
\begin{itemize}
\item JPEG is a C software to compress (\textit{cjpeg}) and uncompress (\textit{djpeg}) full-color and gray-scale JPEG images.
\item MPEG is an optimized implementation for transmitting high quality digital video. The \textit{mpegenc} benchmark is in charge of encode process, and the \textit{mpegdec} benchmark to the decode process.
\item GSM implements an algorithm to reliable voice recognition. The algorithm is based on the implementation of the European GSM 06.10 provisional standard for full-rate speech transcoding. The \textit{gsmenc} benchmark performs the compression frames process, and the \textit{gsmdec} performs the decompression process.
\item \textit{Pegwit} implements the process for public key encryption and authentication. The process uses an elliptic curve over GF($2^{255}$), the codification algorithm SHA1 for hashing, and the symmetric block cipher square. \textit{Pegwitenc} and \textit{pegwitdec} perform encoding and decoding respectively.
\item EPIC (Efficient Pyramid Image Coder) performs image data compression (\textit{epic}) and decompression (\textit{unepic}). The algorithms are based on a biorthogonal critically-sampled dyadic wavelet decomposition and a combined run-length/Huffman entropy coder.
\item ADPCM (Adaptive Differential Pulse Code Modulation) implements speech compression and decompression algorithms. \textit{Rawcaudio} performs the compression while \textit{rawdaudio} performs the decompression. The algorithm they use achieves a compression rate of 4:1 on 16 bits linear PCM samples.
\end{itemize}

Our experiments were carried out in a machine provided with an Intel i5 660 processor at 3.3 GHz. Given that this is a 4-core processor, we run a maximum of 4 experiments at the same time. The available RAM in the machine was 8 Gb, and the operating system was Ubuntu Desktop 14.04.

\subsection{Experimental setup}
As stated before, our experimental framework is parameterized according to the ARM architecture. Then, to evaluate our approach we consider a first level cache memory with a range of sizes between 512B and 64KB, increased in powers of two. With this set of possible cache sizes we cover many of the sizes for some devices of the ARM9 processor family. According to these features, the search space for GE was described in Section \ref{sec:Cache} combining both instruction and data caches. However, the cache characterization is performed for individual memories taking into account only the hardware parameters in the offline characterization process. In our search space, these parameters correspond to $8$ possible cache sizes, $8$ different associativity degrees and $4$ possibles values for block size, as shown in \figurename~\ref{figure:arboles}. Hence, the number of cache characterizations to be run are $256$.

Using our methodology for off-line profiling, we have generated the Mediabench traces using Trimaran tools~\cite{trimaran} in connection with SimpleScalar~\cite{simplescalar} to evaluate the ARM architecture. In the light of the above, the appropriate changes have been made for both SimpleScalar and Trimaran tools to generate application traces able to be processed by the Dinero IV cache simulator.

This kind of experiments require to manage full application traces or execute a number of appropriate instructions which reflects the behavior of the full program. This way prevents to capture a partial or phase behavior. For instance, phase behavior happens at the beginning of the application run, when all data blocks must be copied from the main memory, so all accesses to cache memory are misses. Thus, a small number of instructions can provide results that cannot be considered as significant or conclusive. In this regard, we have simulated every application a maximum of $7.5 \times 10^7$ instructions, or even the full application, if the total number of instructions for the complete execution is lower than this value. The aim is to reach a balance between the simulation time, the size of the program traces and a proper number of instructions.

In addition, GE has been run 10 times for each target application with the aim of reducing the probability to get a local optimum.

Table~\ref{Table:genetic} shows the parameters for GE. We have also applied simple dominance, single point crossover and integer flip mutation operator with probabilities according to recommendations in~\cite{Deb2002}. 
  
\begin{table}
\renewcommand{\arraystretch}{1.3}
\centering
\caption{Parameters for GE.}
\label{Table:genetic}
\begin{tabular}{|c|c|}
\hline
Number of generations & $100$ \\
\hline
Population size & $50$ \\ 
\hline
Probability of crossover & $0.9$ \\ 
\hline
Probability of mutation & $0.01$ \\ 
\hline
\end{tabular}
\end{table}

As stated in (\ref{eq:fitness}) (see Section \ref{GE-phase}), our algorithm requires a baseline cache configuration in order to calculate the goodness of an individual. For these experiments we selected a baseline cache configuration that is similar to the cache of the first core in the GP2X portable game console, which incorporates a dual 200 MHz ARM940T processor.

Table~\ref{table:baselines} shows the size of the cache, the size of the block, the number of ways (associativity degree), replacement policy, search algorithm and, for the data cache, the allocation policy of the baseline configuration.

\begin{table*}
\centering
\renewcommand{\arraystretch}{1.3}
\scriptsize
\caption{Baseline cache configuration. It is formed by parameters of instructions and data caches.}
\label{table:baselines}
\begin{tabular}{|c|c|c|c|c|}
\hline
\multicolumn{5}{|c|}{\textbf{Instructions Cache}} \\
\hline
\hline
\textbf{Size}&\textbf{Block Sz.}&\textbf{Assoc.}&\textbf{Replac.}&\textbf{Search Alg.} \\
\hline
16 KB & 32 B & 4 & LRU & On-Demand \\  \hline
\end{tabular}

\begin{tabular}{c}
\\
\end{tabular}

\begin{tabular}{|c|c|c|c|c|c|}
\hline
\multicolumn{6}{|c|}{\textbf{Data Cache}}\\
\hline
\hline
\textbf{Size}&\textbf{Block Sz.}&\textbf{Assoc.}&\textbf{Replac.}&\textbf{Search Alg.}&\textbf{Alloc.}\\
\hline
16 KB & 32 B & 4 & LRU & On-demand & Copy-Back \\  \hline
\end{tabular}
\end{table*}

Execution time and energy consumption for the baseline cache was computed following the models described in Section \ref{sec:model}.

\subsection{Objective function}

Firstly, we analyze the results regarding the value of the objective function. We would like to recall that objective value $1$ corresponds to the execution time and energy consumption of the baseline cache configuration running each one of the bechmarks. Therefore, we minimize the objective function in order to reduce both features.

Table \ref{tab:obj-function} shows the following statistics, averaged for each one of the benchmarks: objective value, deviation percentage with respect to the best objective found, and number of best results found.

\begin{table*}
\centering
\renewcommand{\arraystretch}{1.3}
\caption{Objective function values.}
\label{tab:obj-function}
\begin{tabular}{|c|c|c|c|}
\hline
Benchmark & Avg. Obj. & Dev. (\%) & \# Best \\
\hline
cjpeg & 0.3986650026 & 0.74\% & 8 \\
djpeg & 0.4746031647 & 3.06\% & 1 \\
epic & 0.3365548961 & 1.24\% & 7 \\
unepic & 0.3977187056 & 0.65\% & 1 \\
gsmdec & 0.3857290316 & 1.15\% & 7 \\
gsmenc & 0.4442631856 & 1.82\% & 4 \\
mpegdec & 0.3682433302 & 1.09\% & 6 \\
mpegenc & 0.3631493163 & 0.89\% & 3 \\
pegwitdec & 0.3663486721 & 0.01\% & 9 \\
pegwitenc & 0.3568943181 & 0.00\% & 10 \\
rawcaudio & 0.3052829963 & 0.00\% & 8 \\
rawdaudio & 0.31533024 & 0.56\% & 9 \\
\hline
 \end{tabular}
\end {table*}

As seen in the table, the average objective values range from $0.47$ to $0.30$, which mean improvements from $53\%$ to $70\%$, with an average improvement of $62\%$ in the objective value. Hence, the optimization obtains really good objective improvements. In addition, the low values of the standar deviation show that GE obtains very close results across the optimization runs. In fact, the number of times the best result is found is higher or equal to $70\%$ in 7 out of the 12 benchmarks. In particular, GE obtains great results in \textit{pegwitdec},  \textit{pegwitenc}, \textit{rawcaudio} and \textit{rawdaudio} benchmarks. On the other hand, despite that GE obtains the best value only once in \textit{djpeg} and \textit{unepic}, the deviation is below $3.1\%$ and $0.7\%$ respectively, which again is a very good result.

These results prove that a metaheuristic like GE, even with a low number of generations and population size, is able to find solutions where the objetive function is reduced above the $62\%$ for the cache optimization problem.

\subsection{Optimization runtime}

We have made another analysus regarding the optimization runtime (RT). Table \ref{tab:runtime} shows the average optimization runtime in hours, the average number of evaluations in relation to the maximum expected for GE, the average runtime for the evaluation of one cache configuration in seconds, the maximum optimization runtime of GE if all evaluations were performed, and the runtime savings of the experiments in relation to that time.

\begin{table*}
\centering
\renewcommand{\arraystretch}{1.3}
\caption{Runtime values. For each benchmark, it shows the average runtime (hours), average number of evaluations vs max. GE, average runtime for one cache evaluation (seconds), maximum runtime for GE (no evaluation recall) and runtime savings in relation to maximum GE.}
\label{tab:runtime}
\begin{tabular}{|c|c|c|c|c|c|}
\hline
Benchmark & RT (h.) & \# Eval. (\%) & Eval. (sec.) & Max GE (h.) & RT Sav. \\ 
\hline
cjpeg & 2.185 & 5.96\% & 26.200 & 36.389 & 93.99\% \\ 
djpeg & 2.647 & 5.74\% & 33.064 & 45.923 & 94.24\% \\ 
epic & 1.405 & 5.87\% & 17.148 & 23.816 & 94.10\% \\ 
unepic & 0.350 & 5.79\% & 4.340 & 6.028 & 94.19\% \\ 
gsmdec & 0.004 & 6.33\% & 0.043 & 0.060 & 93.67\% \\ 
gsmenc & 0.003 & 5.58\% & 0.041 & 0.057 & 94.41\% \\ 
mpegdec & 2.275 & 5.69\% & 28.805 & 40.006 & 94.31\% \\ 
mpegenc & 2.031 & 5.80\% & 25.209 & 35.013 & 94.20\% \\ 
pegwitdec & 0.945 & 6.38\% & 10.667 & 14.815 & 93.62\% \\ 
pegwitenc & 0.869 & 5.98\% & 10.466 & 14.537 & 94.02\% \\ 
rawcaudio & 0.372 & 5.95\% & 4.483 & 6.227 & 94.03\% \\ 
rawdaudio & 0.274 & 5.89\% & 3.343 & 4.642 & 94.10\% \\ 
\hline
 \end{tabular}
\end {table*}

At the same time the optimization was run, we measured the time spent for the evaluation of each cache configuration. As seen in the table, the average evaluation time is high is some of the benchmarks like \textit{djpeg}, \textit{mpegdec} and \textit{mpegenc}. Taking into account this time values, we have calculated the optimization time that a GE execution will spent if the 5000 evaluations were run, that is, if the algorithm will not implement memory for evaluated individuals.

As seen in the savings columns, the execution time is reduced in more than $93.6\%$. Therefore, it is clear that our GE implementation is able to run in reasonable amounts of time due to the memory of evaluated individuals.

The reduced optimization runtimes are caused by two factors. On the one hand, the number of evaluations is low because the exploration of the GE algorithm reaches an area of minimum values making the population to converge to that minimum. In other words, the individuals of the population are becoming similar over the generations. On the other hand, given that our algorithm recalls the evaluation of each individual, the previously evaluated configurations are not sent to the external simulator, but read from the GE memory. This is a very fast process because we implemented it through hashed maps.

\subsection{Execution time and energy consumption}

As stated in \ref{GE-phase}, the objective function incorporates both the execution time and energy consumption values of a given cache configuration. Then, once we have obtained the optimized cache configurations, we show how these results are interpreted in terms of execution time and energy consumption.

\begin{figure*}[t!]
\centering
 \includegraphics[width=0.90\textwidth]{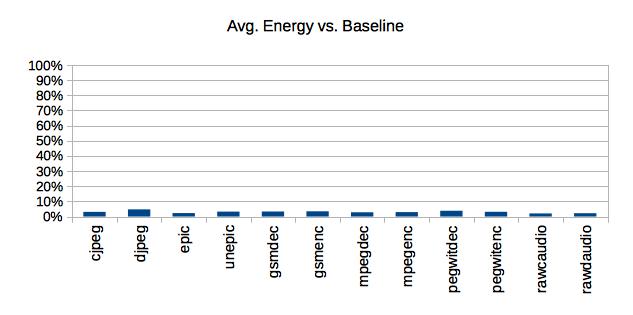}
 \caption{Average energy consumption for the solutions obtained on each benchmark in relation to baseline (100\%).}
  \label{fig:energy}
\end{figure*}

\figurename \ref{fig:energy} plots the average enery consumption of the runs for each benchmark considering that 100\% is the energy consumption of the baseline cache. On average, the solutions we have obtained reach the $3.04\%$ of the baseline. 

\begin{figure*}[t!]
\centering
 \includegraphics[width=0.90\textwidth]{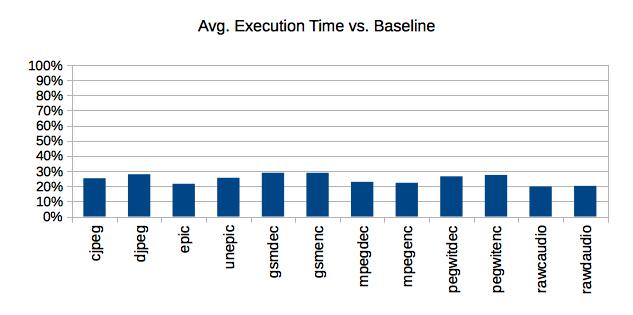}
 \caption{Average execution time for the solutions obtained on each benchmark in relation to baseline (100\%).}
  \label{fig:execTime}
\end{figure*}

\figurename \ref{fig:execTime} plots the average execution time of the runs for each benchmark considering that 100\% is the energy consumption of the baseline cache. On average, the solutions we have obtained reach the $24.74\%$ of the baseline. 

In brief, we have improved the cache behavior by $97\%$ in energy consumption and, at the same time, by $75\%$ in execution time considering a commercial baseline cache. These results were obtained in very short runtimes compared with the expected time that a GE with no evaluation recall will reach.

For the sake of clarity, we would like to remark that the baseline configuration acts as a reference point from which the GE process optimizes. In fact, given that our framework uses the characterization from a cache emulator (Cacti in our case), any baseline configuration could be optimized by providing the framework with the corresponding baseline characterization values.

In addition, given that the benchmark results for the baseline cache are not optimised, the improvements obtained by GE (which are optimised) should be taken as indicative of potential performance gains.

\section{Conclusions and future work}
\label{sec:Conclusions}

The significant increase of multimedia applications in embedded devices such as tablets, smartphones, GPS navigation systems, etc. has caused researchers focus their works on improving either the execution time and energy consumption, or both. In order to improve the execution time, current embedded systems include a cache hierarchy similar to desktop computers. However, the wide range of possible cache configurations complicates the search of the best cache configuration. Not to mention the fact that the execution time and energy consumption are directly affected by the memory access patterns of applications. Therefore, new and efficient techniques have been proposed for years to explore and decide the best cache configuration for different applications. However, we have not found in literature any work that optimize, at the same time, all the parameters that configure both the instructions and the cache memory for a given set of applications.
 
In this paper we have presented an optimization framework based on static profiling and metaheuristics to automatically search for the best cache configuration given a set of applications to be run on a target embedded system. Our optimization scheme considers 11 different parameters that configure both the instruction and the data caches. The framework is divided into two phases. The fist one is an off-line phase in charge to provide the program traces and the features of the set of possible hardware configurations for cache memories. The second phase is an optimization process that applies GE obtaining the cache configuration that would obtain the best behavior for each target application. In order to evaluate a cache configuration, GE obtains the performance of each configuration with the help of the Dinero IV cache simulator. In this way, the grammar of GE eases the communication with Dinero IV given that the phenotype is formed by the parameters and formats required by the simulator call.

In addition, our approach allows to modify the number of parameters to be optimized, as well as the change of the cache simulator by simply modifying the grammar at hand. Besides, if a cache simulator will allow variable length parameters, an appropriate grammar will be able to generate correct simulator calls without any modification in our optimization framework.

We have tested our proposal with twelve applications from the Mediabench benchmarks considering an actual cache configuration as baseline. Our results averaged an improvement of $97\%$ in energy consumption and $75\%$ in execution time versus the baseline cache.

Our future plans include efforts to simplify processes and design an effective framework that allow customize the optimization process for a set of target applications at a time. In addition, we would take into account the different load that memory operations represent on the target applications in order to precisely balance their weight in the optimization. In fact, a natural step forward after this research will be to analyze the influence between execution time and energy results, and consider a multi-objective optimization.

\bibliographystyle{spmpsci}

\bibliography{biblio}

\end{document}